\RequirePackage{fix-cm}
\documentclass[twocolumn]{svjour3}[10pt]       % onecolumn (second format)
\smartqed  % flush right qed marks, e.g. at end of proof
\usepackage{graphicx}

\usepackage{epsfig}
\usepackage{amsmath}
\usepackage{amssymb}
\usepackage{booktabs}
\usepackage{color}
\usepackage{colortbl}
\usepackage{subfigure}
\usepackage{url}
\usepackage[utf8]{inputenc}
\usepackage{fixltx2e}
\usepackage{numprint}
\usepackage{hyphenat}
\usepackage{multirow}
\usepackage{cite}
\definecolor{cwblue1}{rgb}{0.27,0.427,0.623}
\definecolor{cwblue2}{rgb}{0.286,0.454,0.658}
\definecolor{cwblue3}{rgb}{0.733,0.811,0.905}

\newcommand{\NA}{---}
\begin{document}

\title{Are 2D-LSTM really dead for offline text recognition?}



\author{Bastien Moysset         \and
        Ronaldo Messina
}


\institute{Bastien Moysset \at
           A2iA SA, Paris, France \\
           \email{bm@a2ia.com}           %  \\
           \and
           Ronaldo Messina \at
           A2iA SA, Paris, France \\
}

\date{}

\maketitle

\begin{abstract}

There is a recent trend in handwritten text recognition with deep neural networks to replace 2D recurrent layers with 1D,
and in some cases even completely remove the recurrent layers, relying on simple feed-forward convolutional only architectures.
The most used type of recurrent layer is the Long-Short Term Memory (LSTM).
The motivations to do so are many: there are few open-source implementations of 2D-LSTM, even fewer supporting GPU implementations
(currently cuDNN only implements 1D-LSTM); 2D recurrences reduce the amount of computations that can be parallelized,
and thus possibly increase the training/inference time; recurrences create global dependencies with respect to the input,
and sometimes this may not be desirable.

% Nonetheless, recurrent networks have intrinsically more modeling ``power'' compared to feed-forward only nets,
% and this means that if properly regularized, they can yield higher performance. 
Many recent competitions were won by systems that employed networks that use 2D-LSTM layers.
Most previous work that compared 1D or pure feed-forward architectures to 2D recurrent models have done so on simple
datasets or did not fully optimize the ``baseline'' 2D model compared to the challenger model, which was
dully optimized.

In this work, we aim at a fair comparison between 2D and competing models and also extensively evaluate them on
more complex datasets that are more representative of challenging ``real-world'' data, compared
to ``academic'' datasets that are more restricted in their complexity. 
We aim at determining when and why the 1D and 2D recurrent models have different results.
We also compare the results with a language model to assess if linguistic constraints do level the performance
of the different networks.
% By more complexity we mean
% more diversity in styles, contents, layouts, noise and other artifacts. All those aspects
% make it more difficult to attain high performance in text recognition.

Our results show that for challenging datasets, 2D-LSTM networks still seem to provide the
highest performances and we propose a visualization strategy to explain it.

\keywords{Text Line Recognition \and Neural Network \and Recurrent \and 2D-LSTM \and 1D-LSTM \and Convolutional}
\end{abstract}

% 
% 
% \section{Introduction}
% \label{intro}
% Your text comes here. Separate text sections with
% \section{Section title}
% \label{sec:1}
% Text with citations \cite{RefB} and \cite{RefJ}.
% \subsection{Subsection title}
% \label{sec:2}
% as required. Don't forget to give each section
% and subsection a unique label (see Sect.~\ref{sec:1}).
% \paragraph{Paragraph headings} Use paragraph headings as needed.
% \begin{equation}
% a^2+b^2=c^2
% \end{equation}
% 
% % For one-column wide figures use
% \begin{figure}
% % Use the relevant command to insert your figure file.
% % For example, with the graphicx package use
%   \includegraphics{example.eps}
% % figure caption is below the figure
% \caption{Please write your figure caption here}
% \label{fig:1}       % Give a unique label
% \end{figure}
% %
% % For two-column wide figures use
% \begin{figure*}
% % Use the relevant command to insert your figure file.
% % For example, with the graphicx package use
%   \includegraphics[width=0.75\textwidth]{example.eps}
% % figure caption is below the figure
% \caption{Please write your figure caption here}
% \label{fig:2}       % Give a unique label
% \end{figure*}
% %
% % For tables use
% \begin{table}
% % table caption is above the table
% \caption{Please write your table caption here}
% \label{tab:1}       % Give a unique label
% % For LaTeX tables use
% \begin{tabular}{lll}
% \hline\noalign{\smallskip}
% first & second & third  \\
% \noalign{\smallskip}\hline\noalign{\smallskip}
% number & number & number \\
% number & number & number \\
% \noalign{\smallskip}\hline
% \end{tabular}
% \end{table}

\section{Introduction}

Text line recognition is a central piece of most modern document analysis problems. For this reason, many algorithms have been proposed through time to perform this task. The appearance of text images may vary a lot from one image to the other due to background, noises, and especially for handwritten text, writing styles. For this reason modern methods of text recognition tend to use machine learning techniques.

Hidden Markov Models have been used to perform this task with features extracted from the images using Gaussian mixture models \cite{bunke1995off} or neural networks \cite{espana2011improving}.

More recently, like in most of the domains including pattern recognition, deep neural networks have been used to perform this task.

In particular, Graves~\cite{graves2009offline} presented a neural network based on interleaved convolutional and 2D-LSTM (Long-Short Term Memory~\cite{lstm}) layers that were trained using the Connectionist Temporal Classification (CTC) strategy~\cite{ctc}.
This pioneering approach yielded good results on various datasets in several languages \cite{moysset2014} and most of major recent competitions were won by systems with related neural network architectures \cite{bluche2014a2ia,moysset2014,Sanchez2014a,SanchezTRV15,SanchezRTV16,SanchezRTV017}.

Recently, several papers proposed alternative neural network architectures and questioned the need of the 2D-LSTM layers for text recognition systems. Puigcerver et al.~\cite{puigcerver2017multidimensional} propose to use convolutional layers followed by one dimensional LSTM layers to perform handwriting recognition. 
Concurrently, Breuel \cite{breuel2017high} presents a similar architecture for printed text lines while Bluche et al. \cite{bluche2017gated} add convolutional gates as an attention mechanism to this convolutional 1D-LSTM framework.
Borisyuk et al.~\cite{borisyuk2018rosetta}, for the task of scene text recognition, even proposed to completely remove all the LSTM layers
\footnote{Here LSTM denote bi-directional (forward and backward) recurrent layers~\cite{schuster1997bidirectional}, 2D-LSTM introduce top and bottom directions.}
from their network and, thus, to use only convolutional layers.

These techniques that get rid of the 2D-LSTM layers share the same motivation: to be able to use highly parallelizable code increasing the training speed on GPUs. 

%All of them get rid of 2D-LSTMs that were implemented by Garves and used extensively in the precedent years.
%Many recent papers are also not using LSTMs
%We acknowledge that a major reason is that training is faster without 2D-LSTM when using GPU implementations.

Nevertheless, they have in common several trends: the first one is that they evaluate on relatively easy handwritten datasets such as RIMES \cite{rimes} and IAM \cite{iam}, or on machine printed texts. They also tend to use new deep learning techniques and architecture enhancements only for the proposed CNN-1DLSTM models and, therefore, the comparison is not always as fair as it could be.

%But:
%* Easy datasets tend to be used.
%* New deep learning usage and architecture enhancements tend to be used only for proposed CNN-1D-LSTM models.
%* Therefore, the comparison is not always as fair as it could be.

If we do not want to undermine the interest of having competitive networks that enable efficient GPU implementations, we try in this paper to disentangle the advantages and disadvantages of using 2D-LSTMs in text recognition networks.

In this work we propose an extensive analysis of these modern different models and try to give intuitions of when each type of model is the most useful.

We compare the performances of several architectures with :
\begin{itemize}
\item{datasets of various difficulties}
\item{various sizes of datasets}
\item{increasing sizes of networks}
\item{character and word-based language models}
\end{itemize}

We also present some visualization techniques to get intuition of why some architectures work better in some cases.

In section~\ref{sec:models}, we will describe the different architectures that we used. The performance
of the various setups is compared in section~\ref{sec:results} and Section~\ref{sec:visualizations} presents
some visualization to help elucidate how the different models perform with respect to the kind of data that is used.
Finally, Section~\ref{sec:conclusion} discuss the results and present perspectives.
%In section 3, we will compare the system performances in various setups.
%In section 4, we will visualize the activity of the different networks.

%\input{sec_obj}
\section{Models}
\label{sec:models}
All the models used in this paper are designed in order to assess the impact of the choices in the architecture. In particular, we assess the influence of the convolutional gates (Gated Neural Network -- GNN) introduced by Bluche et al. \cite{bluche2017gated}. We also study the impact of 1D-LSTM after the GNN or CNN and the impact of interleaving 2D-LSTM layers between the convolutional layers.

The two recent architectures presented by Bluche et al. \cite{bluche2017gated} and by Puigcerver et al. \cite{puigcerver2017multidimensional} are shown respectively in Tables \ref{tab:architectureCNNpuig} and \ref{tab:architectureGNN}. They are composed of convolutions, plus multiplicative convolutional gates for Bluche et al., followed by 1D-LSTM layers.

\begin{table}
\begin{center}
\caption{Network architecture/hyper-parameters for the CNN-1DLSTM network used by Puigcerver et al. \cite{puigcerver2017multidimensional}.}
\label{tab:architectureCNNpuig}
\resizebox{\linewidth}{!}{%
\begin{tabular}{lcccr}
Layer & Number of & Filter   & Size of the  & Number of  \\ %& feature map size??
      & neurons   & (Stride) & feature maps & parameters \\
\arrayrulecolor{cwblue1} \toprule  
Input & 1 & \NA & $1000\ast128$ & \NA \\
Conv & 16 & $3\times3$ $(1\times1)$ & $1000\ast128$ & 160 \\ %20480000
MaxPooling & 16 & $2\times2$ $(2\times2)$ & $500\ast64$ & \NA \\
Conv & 32 & $3\times3$ $(1\times1)$ & $500\ast64$ & 4\,640 \\ %148480000
MaxPooling & 32 & $2\times2$ $(2\times2)$ & $250\ast32$ & \NA \\
Conv & 48 & $3\times3$ $(1\times1)$ & $250\ast32$ & \numprint{13872} \\ %110976000
MaxPooling & 48 & $2\times2$ $(2\times2)$ & $125\ast16$ & \NA \\
Conv & 64 & $3\times3$ $(2\times2)$ & $125\ast16$ & \numprint{27712} \\ %55424000
Conv & 80 & $3\times3$ $(2\times2)$ & $125\ast16$ & \numprint{46160} \\ %92320000
Tiling & 1280 & $1\times16$ $(1\times16)$ & 125 & \NA \\
1D-LSTM & $2\ast256$ & \NA & 125 & \numprint{3147776} \\ %131072000
1D-LSTM & $2\ast256$ & \NA & 125 &  \numprint{1574912} \\
1D-LSTM & $2\ast256$ & \NA & 125 & \numprint{1574912} \\
1D-LSTM & $2\ast256$ & \NA & 125 &  \numprint{1574912}\\
1D-LSTM & $2\ast256$ & \NA & 125 &  \numprint{1574912}\\
Linear & 110 & \NA & 125 & \numprint{28270} \\
\end{tabular}
}
\end{center}
\end{table}

\begin{table}
\begin{center}
\caption{Network architecture/hyper-parameters for the Gated-CNN-1DLSTM network used by Bluche et al. \cite{bluche2017gated}.}
\label{tab:architectureGNN}
\resizebox{\linewidth}{!}{%
\begin{tabular}{lcccr}
Layer & Number of & Filter   & Size of the  & Number of  \\ %& feature map size??
      & neurons   & (Stride) & feature maps & parameters \\
\arrayrulecolor{cwblue1} \toprule  
Input & 1 & / & $1000\ast128$ & \NA \\
Tiling & 4 & $2\times2$ $(2\times2)$ & $500\ast64$ & \NA \\
Conv & 8 & $3\times3$ $(1\times1)$ & $500\ast64$ & 296 \\ %9472000
Conv & 16 & $2\times4$ $(2\times4)$ & $250\ast16$ & 1040 \\ %4160000
GatedConv & 16 & $3\times3$ $(1\times1)$ & $250\ast16$ & 2320 \\ %9280000
Conv & 32 & $3\times3$ $(1\times1)$ & $250\ast16$ & 4640 \\ %18560000
GatedConv & 32 & $3\times3$ $(1\times1)$ & $250\ast16$ & 9248 \\ %36992000
Conv & 64 & $2\times4$ $(2\times4)$ & $125\ast4$ & 16448 \\ %8192000
GatedConv & 64 & $3\times3$ $(1\times1)$ & $125\ast4$ & 36928 \\ %18464000
Conv & 128 & $3\times3$ $(1\times1)$ & $125\ast4$ & 73856 \\ %36928000
MaxPooling & 128 & $1\times4$ $(1\times4)$ & 125 & \NA \\
1D-LSTM & $2\ast128$ & / & 125 & 263168 \\ %32768000
Linear & $2\ast128$ & / & 125 & 32768 \\ %4128000
1D-LSTM & $2\ast128$ & / & 125 & 263168 \\ %32768000
Linear & $2\ast110$ & / & 125 & 28380 \\ %3547500
\end{tabular}
}
\end{center}
\end{table}

We also propose new architectures that take inspiration from the GNN-1DLSTM architecture. We create a CNN-1DLSTM architecture similar to the GNN-1DLSTM by removing the gated multiplicative convolutions. We also create CNN- and GNN-only architectures by removing the final 1D-LSTM layers from the GNN-1DLSTM and CNN-1DLSTM models.

Finally, we propose a network with interleaved convolutional and 2D-LSTM layers by replacing each couple of a convolutional and a multiplicative gate layers from the Bluche et al. GNN-1DLSTM network by a 2D-LSTM layer. For simplicity, this model will be called 2DLSTM through this paper, even if it also includes convolutional and 1D-LSTM layers. 

The architecture is presented in Table \ref{tab:architecture2D-LSTM}. This model has approximately the same number of parameters than the GNN-1DLSTM model from Bluche et al. and only a bit higher ($\times 1.5$) number of operations is needed as illustrated in Table \ref{tab:archiParams}. 

We also observe in Table \ref{tab:archiParams} that the Puigcerver architecture is significantly bigger in term of operations and number of parameters. Indeed, the number of parameters is more than 11 times higher than in the proposed architecture and the number of operations is almost 5 times higher. For this reason, we propose a larger 2DLSTM architecture by multiplying the depth of all the feature maps by 2. This architecture, called 2DLSTM-X2, has still a significantly smaller number of parameters than the Puigcerver architecture, and fewer operations.  

For all of these models, no tuning of the filter sizes and layer depths was performed by us, on any datasets.  This, in order not to bias our experiments by improving one model more than the others. 

\begin{table}
\begin{center}
\caption{Network architecture/hyper-parameters for the 2DLSTM network presented in this paper.}
\label{tab:architecture2D-LSTM}
\resizebox{\linewidth}{!}{%
\begin{tabular}{lcccr}
Layer & Number of & Filter   & Size of the  & Number of  \\ %& feature map size??
      & neurons   & (Stride) & feature maps & parameters \\
\arrayrulecolor{cwblue1} \toprule  
Input & 1 & / & $1000\ast128$ & \NA \\
Tiling & 4 & $2\times2$ $(2\times2)$ & $500\ast64$ & \NA \\
Conv & 8 & $3\times3$ $(1\times1)$ & $500\ast64$ & 296 \\ %9472000
2D-LSTM & $4\ast8$ & $1\times1$ $(1\times1)$ & $500\ast64$ & 2560 \\ %81920000
Conv & $4\ast16$ & $2\times4$ $(2\times4)$ & $250\ast16$ & 4160 \\ %16640000
2D-LSTM & $4\ast20$ & $1\times1$ $(1\times1)$ & $250\ast16$ & 22800 \\ 
Conv & $4\ast32$ & $2\times4$ $(2\times4)$ & $125\ast4$ & 20480 \\
2D-LSTM & $4\ast32$ & $1\times1$ $(1\times1)$ & $125\ast4$ & 90400 \\
Conv & $4\ast40$ & $2\times4$ $(2\times4)$ & $63\ast1$ & 81920 \\
Conv & 128 & $3\times1$ $(1\times1)$ & 63 & 24704 \\
1D-LSTM & $2\ast128$ & / & 63 & 263168 \\
Linear & $2\ast128$ & / & 63 & 32768 \\
1D-LSTM & $2\ast128$ & / & 63 & 263168 \\
Linear & $2\ast110$ & / & 63 & 28380 \\
\end{tabular}
}
\end{center}
\end{table}

\begin{table}
\begin{center}
\caption{Comparison of the number of parameters and operations needed for the different architectures. For the number of operations, an image of size $128 \times 1000$ $(H \times W)$ is considered.}
\label{tab:archiParams}
\resizebox{\linewidth}{!}{%
\begin{tabular}{lcc}
Architecture & Number of & Number of  \\ %& feature map size??
             & parameters & operations \\
\arrayrulecolor{cwblue1} \toprule  
CNN-1DLSTM Puigcerver et al. & 9.6M & 1609M \\
GNN-1DLSTM Bluche et al. & 799k & 216M \\
2DLSTM & 836k & 344M \\
2DLSTM-X2 & 3.3M & 1340M \\
\end{tabular}
}
\end{center}
\end{table}

\subsection{Language models}
\label{sec:languagemodel}
Having recurrent layers at the output of the network might cause some language-related information
to be used by the optimizer during training, because the order of the labels presented is in some
ways predictable. It can be seen as a ``latent'' language model. Therefore, we also evaluate the different models with the aid of an ``external'' language model (LM).

It is straight forward to use a weighted finite-state transducer (FST) representation of a LM~\cite{FST} to apply
syntactic and lexical constraints to the posterior probabilities predicted by the neural networks
as shown in~\cite{moysset2014} (we estimate priors for each character from the training data and a value of $0.7$ for
the weight given to those priors); we omit here the non-essential details of interfacing neural network outputs and FSTs. 
Pruning is used to reduce the size of the LMs, no effort was done in order to optimize the LMs, as that was not
the aim of this experience. The SRI~\cite{SRI} toolkit is used in the construction of all LMs and the
Kaldi~\cite{KALDI} decoder is used to obtain the 1-best hypothesis.

We use the text from wikipedia dumps\footnote{https://dumps.wikimedia.org/frwiki/20180701/frwiki-20180701-pages-articles-multistream.xml.bz2}~\footnote{ttps://dumps.wikimedia.org/enwiki/20181011/enwiki-20181001-pages-articles-multistream.xml.bz2} to estimate word and character-level language models
for French and English models; for the READ (see Section~\ref{sec:datasets}) we just used the training data. As it is not in modern German we cannot rely on wikipedia for textual data.
In the character-level LMs, we add the space separating words as a valid token (it is also predicted by the
neural network). In text recognition LMs, punctuation symbols are considered as tokens. We split numbers 
into digits to simplify the model.
Some characters were replaced by the most similar that is modeled (e.g. the ligature ``œ'' is replaced by ``oe'', ’~\footnote{Right Single Quotation Mark} is replaced by a single quote, \textit{en} and \textit{em} dashes by a single dash, etc.)
Lines containing characters that are not modeled are ignored, and some ill-formed lines that
could not be parsed are also ignored. 
%The resulting data amount to \numprint{13949232} lines, and \numprint{2621753} word tokens (punctuation included).

The sizes of the different evaluation sets are given in Table~\ref{tab:evalsizes}, in terms of total number
of tokens and the cardinality of that set.

\begin{table}
\begin{center}
\caption{Composition of the different evaluation sets.}
\label{tab:evalsizes}
%\resizebox{\linewidth}{!}{%
\begin{tabular}{lcc}
Dataset & Words & Vocab
 \\ \hline
Maurdor-HWR-Dev2.1 & 10780 & 2363
 \\
Maurdor-HWR-Dev2.2 & 11211 & 2416
 \\
Maurdor-PRN-Dev2.1 & 66410 & 7128 \\
Maurdor-PRN-Dev2.2 & 51521 & 7003
 \\
RIMES-valid & 7839 & 1347
 \\
RIMES-test & 7411 & 1285 \\
MaurdorDev-TiersSimple-ForGNN & 3297 & 690
 \\
MaurdorDev-TiersDur-ForGNN & 2969 & 1032
 \\
linesWithLessThan8Letters & 1388 & 252
 \\
linesWithBetween8And19Letters & 3542 & 697 \\
linesWithMoreThan19Letters & 6281 & 1779 \\ \hline
IAM-valid & 9475 & 2429
 \\
IAM-test & 27095 & 5200 \\ \hline
READ-valid & 8414 & 1491 \\
\end{tabular}
%}
\end{center}
\end{table}

From the data, word 3-gram language models with different number of tokens in the vocabulary,
ranging in 25k, 50k, 75k, 100k, and 200k are estimated for EN and FR models, for READ the vocabulary was quite small (less than
7k words) so no limitation was imposed; words out of vocabulary (OOV) can not be recognized, and we present
the number of those words for the different evaluation datasets in Table~\ref{tab:oov}. The OOV ratio for the READ validation set is 14.8\%.
We also estimate character $n$-grams, where $n=5,6,7$; there are no OOVs in this case, and practically all characters in evaluation data are modeled. 
The number of characters in the test datasets that was not modeled is too small and should not have influence in the CER.
Given the sizes of the databases from wikipedia, we could probably go further than a word 3-gram, but we are \textit{not} interested in maximizing performance, just assess the impact of a LM on the decoding results. 

\begin{table}
\begin{center}
\caption{OOV ratios for the different evaluation sets.}
\label{tab:oov}
\resizebox{\linewidth}{!}{%
\begin{tabular}{lccccc}
\multirow{2}{*}{Dataset} & \multicolumn{5}{c}{LM-vocabulary} \\
 &  25k & 50k & 75k & 100k & 200k
 \\ \hline
Maurdor-HWR-Dev2.1 & 11.22 & 8.67 & 7.54 & 6.78 & 5.36
 \\ 
Maurdor-HWR-Dev2.2 & 10.29 & 7.69 & 6.68 & 5.88 & 4.75
 \\ 
Maurdor-PRN-Dev2.1 & 10.97 & 8.17 & 7.29 & 6.54 & 5.15
 \\ 
Maurdor-PRN-Dev2.2 & 10.95 & 7.91 & 6.8 & 6.05 & 4.59
 \\ 
RIMES-valid & 8.06 & 5.43 & 4.59 & 2.74 & 1.8
 \\ 
RIMES-test & 8.88 & 6.4 & 5.37 & 3.56 & 2.59 \\ 
MaurdorDev-TiersSimple-ForGNN & 9.65 & 7.22 & 6.19 & 5.73 & 4.94
 \\
MaurdorDev-TiersDur-ForGNN & 14.11 & 11.42 & 10.27 & 9.53 & 7.48
 \\
linesWithLessThan8Letters & 10.09 & 8.43 & 7.06 & 6.56 & 5.40
 \\
linesWithBetween8And19Letters & 11.46 & 9.06 & 8.44 & 7.76 & 6.52
 \\
linesWithMoreThan19Letters & 9.68 & 6.75 & 5.60 & 4.66 & 3.61 \\ \hline
IAM-valid & 6.22 & 3.85 & 2.60 & 2.17 & 1.44
 \\
IAM-test & 6.57 & 3.77 & 2.79 & 2.15 & 1.24 \\ 
\end{tabular}
}
\end{center}
\end{table}
\section{Experimental results}
\label{sec:results}

In this section, we show and analyze the results of the different neural network architectures proposed in Section \ref{sec:models}, in various setups.
In section \ref{sec:expeSetUp}, we describe the experimental setup and detail the datasets that we use. Then, we analyze the results on various datasets in section \ref{sec:expeDataset}, with various hyper-parameter choices in section \ref{sec:expeHyperparameters}. We analyze the robustness of the methods to dataset transfer in section \ref{sec:expeRobustness} and we study the impact of language models in section \ref{sec:expeLM}.

\subsection{Experimental setup}
\label{sec:expeSetUp}

\subsubsection{Datasets}
\label{sec:datasets}

Five different datasets are used in our experiments. RIMES \cite{rimes} and IAM \cite{iam} are traditional handwritten text datasets in respectively French and English. Because there is no background, no variation in scanning procedure and because the segmentation is made by hand, they can be considered as easy handwritten datasets.
The READ dataset \cite{SanchezRTV017} comprises historical handwritten images. Even if the documents are written by only a few number of writers, the background noises related to the age of the documents make it more difficult.
Finally, the MAURDOR dataset \cite{LneICASSP2014,maurdor} includes modern documents of several types (letters, forms, etc), from several sources, and several scanning methods. This variety, and the fact that data are annotated automatically makes it a challenging dataset. We experimented on two subsets of the MAURDOR dataset: the handwritten and the machine printed text lines, both in French.
%on the one side and the printed text lines in French on the other side.

\begin{figure*}[t]
   \centering
   \begin{minipage}[b]{0.40\linewidth}
      \centering
      \includegraphics[height=0.4cm]{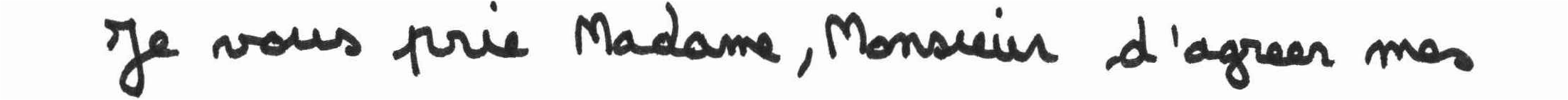}
      \includegraphics[height=0.4cm]{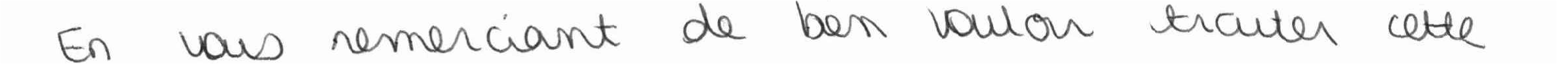}
      \includegraphics[height=0.4cm]{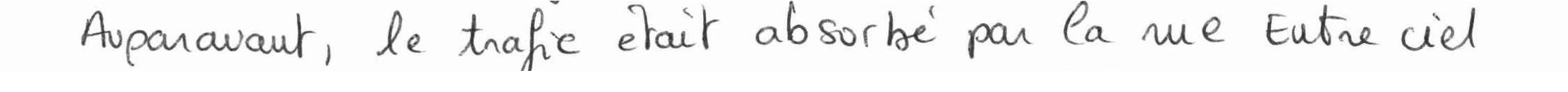}
      %\vspace{1px}
      \\
      Rimes
   \end{minipage}
   \begin{minipage}[b]{0.40\linewidth}
      \centering
      \includegraphics[height=0.4cm]{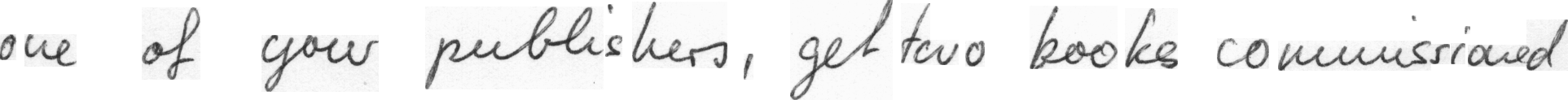}
      \includegraphics[height=0.4cm]{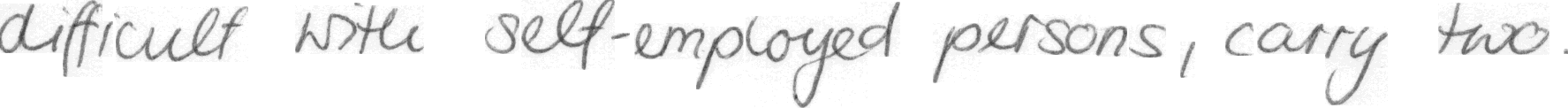}
      \includegraphics[height=0.4cm]{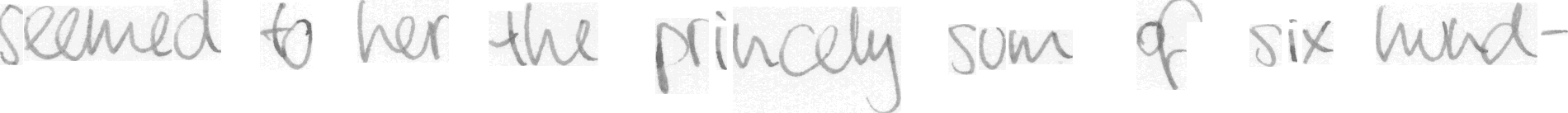}
      %\vspace{1px}
      \\
      IAM
   \end{minipage}

   \vspace{0.2cm}
   
   \begin{minipage}[b]{0.3\linewidth}
      \centering
      \includegraphics[height=0.6cm]{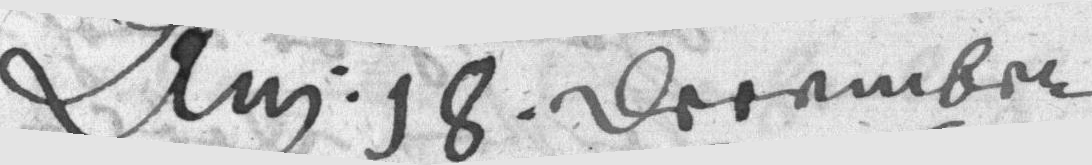}
      \includegraphics[height=0.6cm]{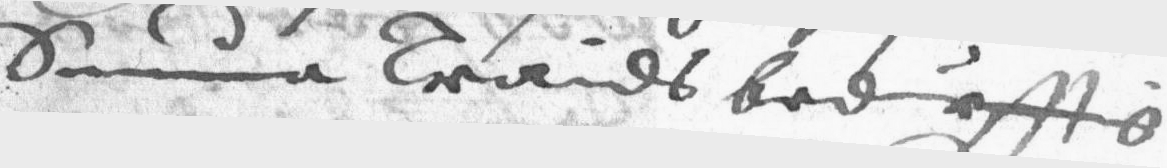}
      \includegraphics[height=0.5cm]{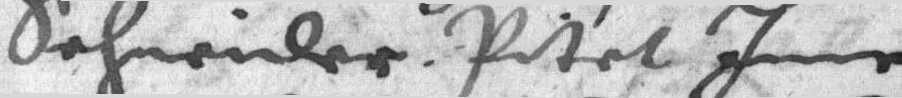}
      %\vspace{1px}
      \\
      Read
   \end{minipage}
   \begin{minipage}[b]{0.30\linewidth}
      \centering
      \includegraphics[height=0.4cm]{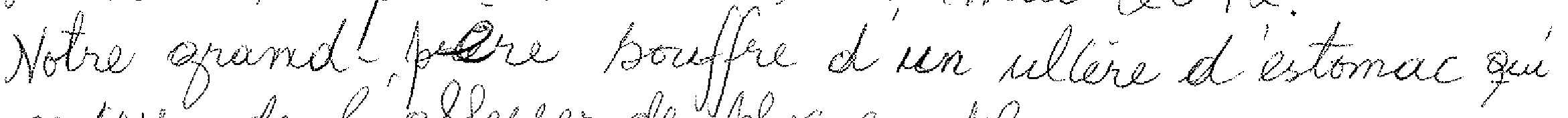}
      \includegraphics[height=0.6cm]{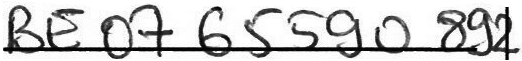}
      \includegraphics[height=0.7cm]{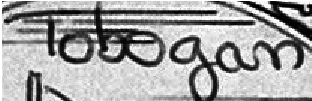}
      %\vspace{0.1cm}
      \\
      Maurdor Handwritten
   \end{minipage}
   \begin{minipage}[b]{0.30\linewidth}
      \centering
      \includegraphics[height=0.4cm]{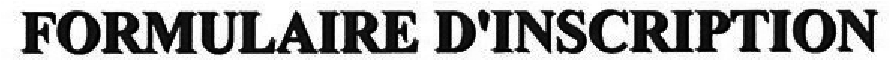}
      \includegraphics[height=0.4cm]{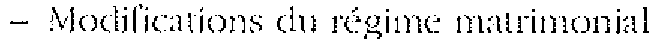}
      \includegraphics[height=0.7cm]{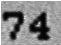}
      %\vspace{0.1cm}
      \\
      Maurdor Printed
   \end{minipage}

   \caption{Samples of text lines from the five datasets used in this work.}
   \label{fig:illustrationsLignesDataset}
\end{figure*}

The details of the dataset compositions are given in Table \ref{tab:datasetSizes} and some examples of text lines from each dataset can be found in Figure~\ref{fig:illustrationsLignesDataset}. These dataset samples illustrate the varying difficulty of the text recognition task from one dataset to the other. In particular, we can observe the heterogeneity of the Maurdor handwritten dataset and the noisiness of the READ historical documents.

\begin{table}
\begin{center}
\caption{Comparison of the number of text lines in the different used datasets.}
\label{tab:datasetSizes}
\resizebox{\linewidth}{!}{%
\begin{tabular}{lcccccc}
Dataset & Train lines & Valid lines & Test lines \\
RIMES & 10532 & 801 & 778  \\
IAM  & 6482 & 976 & 2915  \\
MAURDOR Handwritten & 26870 & 2054 & 2035  \\
MAURDOR Printed  & 97729 & 9182 & 7899  \\
READ & 16734 & 2086 & /  \\
\end{tabular}
}
\end{center}
\end{table}

Note that for MAURDOR, printed and handwritten, we used the technique described in \cite{munchausen} to extract the line images and associated labels from the given paragraph-level annotations.
For all datasets, traditional training, development and test sets are used.

\subsubsection{Training of the networks}

If not otherwise stated, the following parameters are used to train all our different networks. No tuning of these parameters have been perform and they mostly have been taken from Bluche et al. \cite{bluche2017gated}.

All the networks are trained from scratch which means that no pre-training or transfer from another dataset is used. For all the trainings, we use a RmsProp \cite{tielmann} optimizer with an initial learning rate of 0.0001 and mini-batches of size 8. Early stopping is applied after 20 non-improving epochs. We use Glorot \cite{glorot2010understanding} initialization for all the free parameters of our networks. Dropout \cite{pham2014dropout,SriHin14Dropout} is applied after several convolutional layers of all the networks with a probability of 0.5.

In Puigcerver et al. \cite{puigcerver2017multidimensional}, it is proposed to train the network with batch normalization and data augmentation. To ease the comparison with other networks, we have not used both of these training tricks in our experiments.

Finally, for all the networks, input images are all isotropically rescaled to a fixed height of 128 pixels, they are converted to gray-scale and normalized in mean and standard deviation.

\subsection{Results with respect to the dataset.}
\label{sec:expeDataset}

In this section, we compare the behavior of the networks described in Section \ref{sec:models} on several datasets. Once again, hyper-parameters have not been tuned for either of the networks or datasets. 

In Table \ref{tab:expeDatasetDev} and Table \ref{tab:expeDatasetEval}, we show results on the five datasets described in section \ref{sec:datasets}, respectively for the validation and the test sets. Both character error rates (CER) and word error rates (WER) are computed with the Levenshtein distance \cite{levenshtein}. 

The first observation is that for all the datasets, having LSTM layers on top of the layer is essential the CNN and GNN-only networks show deteriorated performances in all setups. 

Between the models having recurrences in their decoder, we observe that for easier datasets like MAURDOR (machine printed), all models achieve similar performances. On the contrary, for more difficult datasets, like the handwritten subset of MAURDOR, or the READ dataset, results are more varying. 

We see that for similar size networks, the networks having 2D-LSTMs tend to perform better than those without. One can suppose that this is due to the presence of noise in the images coming from image deterioration in the case of READ and from the unconstrained layout and the presence of parts of other lines in the case of Maurdor. Having 2D-LSTM layers on low-level layers enable to convey and learn the spatial context information useful for ignoring these noises.

\begin{table*}
\begin{center}
\caption{Text recognition results (CER,WER) of various models on the validation sets of the five chosen datasets. No language model applied.}
\label{tab:expeDatasetDev}
\resizebox{\linewidth}{!}{%
\begin{tabular}{lccccccc}
 & CNN & GNN & CNN-1DLSTM & GNN-1DLSTM & 2DLSTM & 2DLSTM-X2 & CNN-1DLSTM Puigcerver \\
RIMES & 16.57\% , 56.55\% & 11.71\% , 43.49\%  & 4.27\% , 16.59\% & 4.00\% , 15.91\% & 3.32\% , 13.24\% & 3.14\% , 12.48\% & 2.90\% , 11.68\% \\
IAM & 18.53\% , 60.59\% & 85.65\% , 99.38\% & 7.15\% , 25.83\% & 6.18\% , 23.11\% & 5.41\% , 20.15\% & 5.40\% , 20.40\% & 4.62\% , 17.31\% \\
MAURDOR Handwritten & 32.40\% , 80.07\% & 23.40\% , 66.91\% & 13.30\% , 42.25\% & 11.80\% , 38.85\% & 10.10\% , 33.86\% & 8.81\% , 31.35\% & 9.24\% , 54.28\% \\ 
MAURDOR Printed FR & 5.74\% , 20.20\% & 3.66\% , 13.57\% & 1.67\% , 5.55\% & 1.58\% , 5.32\% & 1.59\% , 5.21\% & 1.59\% ,5.27\% & 1.64\% , 5.37\% \\
Historical ICFHR 2016 READ & 26.25\% , 78.84\% & 18.28\% , 63.56\% & 12.39\% , 41.33\% & 10.40\% , 37.61\% & 7.97\% , 29.48\% & 7.71\% , 29.31\% & 7.83\% , 29.46\% \\
\end{tabular}
}
\end{center}
\end{table*}

\begin{table*}
\begin{center}
\caption{Text recognition results (CER,WER) of various models on the test sets of the four chosen datasets. No language model applied.}
\label{tab:expeDatasetEval}
\resizebox{\linewidth}{!}{%
\begin{tabular}{lccccccc}
 & CNN & GNN & CNN-1DLSTM & GNN-1DLSTM & 2DLSTM & 2DLSTM-X2 & CNN-1DLSTM Puigcerver \\
RIMES & 20.21\% , 63.33\% & 14.61\% , 51.14\% & 6.14\% , 20.11\% & 5.75\% , 19.74\% & 4.94\% , 16.03\% & 4.80\% , 16.42\% & 4.39\% , 14.05\% \\
IAM & 23.13\% , 65.44\% & 85.66\% , 99.55\% & 11.52\% , 35.64\% & 10.17\% , 32.88\% & 8.88\% , 29.15\% & 8.86\% , 29.31\% & 7.73\% , 25.22\% \\
MAURDOR HWR FR & 33.64\% , 82.40\% & 24.61\% , 69.67\% & 14.29\% , 44.43\% & 12.58\% , 40.74\% & 10.80\% , 37.12\% & 9.36\% , 33.75\% & 9.80\% , 56.66\% \\ 
MAURDOR PRN FR & 6.35\% , 19.98\% & 4.07\% , 13.23\% & 2.01\% , 5.86\% & 1.81\% , 5.43\% & 1.94\% , 5.51\% & 1.84\% , 5.39\% & 1.91\% , 5.75\% \\
\end{tabular}
}
\end{center}
\end{table*}

After having compared the performances of different architecture on datasets from different sources, we will study the dynamics of the results when selecting some sub-datasets from the main source, namely the Maurdor handwritten dataset.

First, we compare the impact of the number of labeled text lines available. Results are shown in Table \ref{tab:expeNumberOfLines}. For several architectures (CNN, GNN-1DLSTM, 2DLSTM), we train networks with a various number of training text lines.

Text lines are chosen randomly from the MAURDOR handwritten training set and we compose 4 sets of \numprint{2000}, \numprint{4000}, \numprint{8000} and \numprint{16000} text lines that we compare with the full training set of about \numprint{24000} text lines.

We observe that the more data is available for training, the more the network with 2D-LSTM layers in the encoder outperforms the GNN-1DLSTM model that only has gates. Similarly, we observe that the convolutional-only network is less outperformed for very small datasets. For the small datasets containing \numprint{2000} and \numprint{4000} lines, the 2D-LSTM model is outperformed by the GNN-1DLSTM model. This can be explained by the fact that the 2D-LSTM network is more powerful in the sense that it can convey contextual information and thus is both more prone to overfitting when the number of data is low and enable to benefit more from the increase in data amount.

\begin{table}
\begin{center}
\caption{Text recognition results (CER,WER) on the MAURDOR Handwritten validation set with varying amount of training data and several network architectures.}
\label{tab:expeNumberOfLines}
\resizebox{\linewidth}{!}{%
\begin{tabular}{lccc}
\#lines & CNN & GNN-1DLSTM & 2DLSTM \\ 
\numprint{2000}  & 51.64\% , 93.63\% & 40.70\% , 81.43\% & 69.74\% , 98.78\% \\
\numprint{4000}  & 45.42\% , 90.47\% & 32.37\% , 73.05\% & 33.85\% , 74.62\% \\
\numprint{8000}  & 40.81\% , 87.09\% & 24.34\% , 62.30\% & 21.25\% , 57.62\% \\
\numprint{16000} & 32.92\% , 79.32\% & 14.18\% , 44.33\% & 12.51\% , 40.76\% \\
All   & 32.40\% , 80.07\% & 11.80\% , 38.85\% & 10.1\% , 33.9\% \\
\end{tabular}
}
\end{center}
\end{table}

Secondly, we split our MAURDOR handwritten training set in three subsets based on the character error rate of each line. We create an easy set made of the 8\,000 line images with the lowest error rates, and a hard set made of the 8\,000 line images with the highest error rates. We compare these two sets with a third set composed of 8\,000 randomly chosen text lines. Networks are trained on each of these subsets and they are evaluated on three subsets of the validation set composed with a similar method.

The results of this data hardness experiments are shown in Table \ref{tab:expeEasyHard}. If the 2DLSTM based network outperforms the GNN-1DLSTM network for all the training and testing combinations, we do not observe significant variation of performance shift with respect to the hardness of the line to be recognized. We suspect that this is related to the hardness selection process which is biased to select smaller sequences in the harder set (because only a couple of errors will create a high proportion of character error rate) and that these sequences are less likely to use the long term context information that the 2DLSTMs can convey.

\begin{table*}
\begin{center}
\caption{Text recognition results (CER,WER) on the MAURDOR Handwritten validation set with networks trained on data selected function of their hardness to be recognized.}
\label{tab:expeEasyHard}
\resizebox{\textwidth}{!}{%
  \begin{tabular}{lcccccc}
  & Easy set & Easy set & Random set & Random set & Hard set & Hard set \\
  & GNN-1DLSTM & 2DLSTM & GNN-1DLSTM & 2DLSTM & GNN-1DLSTM & 2DLSTM \\
  Easy 8\,000 & 6.5\% , 24.7\% & 5.6\% , 23.0\% & 21.5\% , 55.4\% & 19.1\% , 52.2\% & 42.4\% , 84.7\% & 38.4\% , 81.7\%  \\  
  Random 8\,000 & 9.4\% , 34.6\% & 7.6\% , 31.6\% & 24.3\% , 62.3\% & 21.2\% , 57.6\% & 43.0\% , 87.7\% & 38.9\% , 83.5\%  \\
  Hard 8\,000 & 9.6\% , 43.7\% & 7.7\% , 37.6\% & 16.8\% , 52.8\% & 14.7\% , 47.9\% & 33.2\% , 79.1\% & 29.8\%, 73.3\%  \\
  \end{tabular}
}
\end{center}
\end{table*}

Thirdly, we compare in Table \ref{tab:expeSizeOfLines} the performances of the different networks, all trained on the whole Maurdor Handwritten French training dataset, on three subsets of the validation set of comparable size. The lines are distributed in the three created datasets with respect to the number of characters they contain. The first one is made with all the lines that have less than 8 characters, the second with lines that have between 8 and 19 characters and the last with lines of more than 19 characters.

We observe that the ranking of the networks remain the same whatever the subset we evaluate on. At character level, the longest lines are easier to recognize. We do not observe significant changes in the score dynamics between the model when the size of the lines changes.

\begin{table*}
\begin{center}
\caption{Text recognition results (CER,WER) on three subsets of the MAURDOR Handwritten validation set made of lines with varying number of characters with several networks trained on the whole training set. No language model is used.}
\label{tab:expeSizeOfLines}
\resizebox{\linewidth}{!}{%
\begin{tabular}{lccccc}
 & CNN & GNN & CNN-1DLSTM & GNN-1DLSTM & 2DLSTM \\
Short lines (less than 8 char) & 48.9\% , 89.9\% & 37.4\% , 76.3\% & 19.9\% , 45.0\% & 17.1\% , 40.4\% & 15.0\% , 35.4\% \\
Medium lines (between 8 and 19 chars) & 39.7\% , 98.1\% & 28.4\% , 83.6\% & 15.3\% , 53.7\% & 13.9\% , 49.9\% & 11.5\% , 45.4\% \\
Long lines (more than 19 chars) & 30.2\% , 76.4\% & 22.1\% , 64.3\% & 13.4\% , 41.4\% & 11.7\% , 37.9\% & 10.2\% , 34.7\% \\
\end{tabular}
}
\end{center}
\end{table*}

In summary, we have observed an inter-dataset correlation that the 2D-LSTMs help more to improve the results over the GNN-1DLSTM system when the lines are noisier and harder but we have not been able to identify intra-dataset differences when taking longer or higher CER text lines.

\subsection{Results with respect to the chosen architecture hyper-parameters}
\label{sec:expeHyperparameters}

We then compare the impact of some architectural choices on the performance of our two main networks, GNN-1DLSTM and 2DLSTM. 

First, we study the impact on the recognition performances of the number of layer in the network. For both networks, we add/remove extra convolutions of filter sizes $3 \times 3$ and of stride 1 in the encoder part of our network. Results are shown in Table \ref{tab:expeArchitectureEncoder}. We can see that the increase of the number of layers slightly benefits the 2DLSTM model while it does not impact much the GNN-1DLSTM one. 

\begin{table}
\begin{center}
\caption{Comparison of GNN-1DLSTM and 2DLSTM models for a varying number of layers in the encoder. Results on the French handwritten validation set, no language model (CER, WER).}
\label{tab:expeArchitectureEncoder}
% \resizebox{\linewidth}{!}{%
\begin{tabular}{lcc}
 & GNN-1DLSTM & 2DLSTM \\
6 layers & 12.42\% , 39.62\ & 11.14\% , 37.52\% \\
8 layers (Ref) & 11.80\% , 38.85\% & 10.10\% , 33.86\% \\
12 layers & 12.07\% , 38.69\% & 9.57\% , 33.02\% \\
\end{tabular}
% }
\end{center}
\end{table}

Then, we change the number of bidirectional 1D-LSTM layer in the decoder of the networks; reducing it to 1 and extending it to 5. As stated in Table \ref{tab:expeArchitectureDecoder}, for both model, the performance is positively correlated to the number of 1D-LSTM layers in the decoder. But the impact of this increase is more important for the GNN-1DLSTM model. Probably because the 2D-LSTMs models can already take advantage of the horizontal context information transmission of their 2D-LSTM layers if needed.

\begin{table}
\begin{center}
\caption{Comparison of GNN-1DLSTM and 2DLSTM models for a varying number of 1D-LSTM layers in the decoder. Results on the French handwritten validation set, no language model (CER, WER).}
\label{tab:expeArchitectureDecoder}
% \resizebox{\linewidth}{!}{%
\begin{tabular}{lcc}
 & GNN-1DLSTM & GR-2DLSTM \\
1 1DLSTM & 14.66\% , 46.12\% & 11.57\% , 38.6\% \\
2 1DLSTM (Ref) & 11.80\% , 38.85\% & 10.10\% , 33.86\%   \\
5 1DLSTM & 9.38\% , 31.69\% & 9.30\% , 31.68\% \\
\end{tabular}
% }
\end{center}
\end{table}

One of the key differences between the GNN-1DLSTM architecture of Bluche and al. and the CNN-1DLSTM architecture of Puigcerver and al., apart from the size of the networks, resides in the way the 2D-signal is collapsed to a 1D-signal. In Bluche et al., and in our 2DLSTM network presented through this paper and inspired from it, a max-pooling is used between the vertical locations while in Puigcerver et al, the features of all the 16 vertical locations are concatenated. 

This concatenation method causes a large increase in the number of parameters of the network because the number of parameters of this interface layer (the first 1D-LSTM layer) is multiplied by the number of vertical positions. But we wanted to assess its influence on the other models. We can see in Table \ref{tab:expeArchitecturePooling} that this change of interface function, and the associated increase in parameters, has no significant impact on the GNN-1DLSTM model, while it improves the 2DLSTM one, especially for the character error rate. 

\begin{table}
\begin{center}
\caption{Comparison of GNN-1DLSTM and 2DLSTM models when using a concatenation of the features from all the vertical positions instead of a max-pooling at the interfaces between the encoders and the decoders. Results on the French handwritten validation set, no language model (CER, WER). }
\label{tab:expeArchitecturePooling}
% \resizebox{\linewidth}{!}{%
\begin{tabular}{lcc}
 & GNN-1DLSTM  & 2D-LSTM \\
Max-Pooling (Ref) & 11.80\% , 38.85\% & 10.10\% , 33.86\% \\
 & 799k parameters & 836k parameters \\
Concatenation & 11.73\% , 39.16\% & 8.93\% , 31.90\% \\
 & 1192k parameters & 1289k parameters \\
\end{tabular}
% }
\end{center}
\end{table}

We also wanted to assess how the increase of the depths of the feature maps affect the results. For this reason, we multiplied all these depths by 2 for both the GNN-1DLSTM and the 2DLSTM models. By doing that, we approximately multiply the number of parameters of the networks and the number of operations needed to process a given image by 4 as illustrated in Table \ref{tab:archiParams} and get closer to the size of the Puigcerver CNN-1DLSTM network though still about three times smaller. 
We also tried to split the depths of these feature maps by 2 and 4. We compare it with the reference feature map depths detailed in Tables \ref{tab:architectureGNN} and \ref{tab:architecture2D-LSTM}.

As shown in Table \ref{tab:expeArchitectureParams}, both the architectures benefit from this increase in feature map depths. A higher increase occurs for the 2DLSTM network that behaves worse than the GNN-1DLSTM network for very small depths (divided by 4), similarly for medium sized depths (divided by 2 or the reference ones) and significantly better for deep feature maps (multiplied by 2). This is probably due to the fact that more feature maps enable to learn more various information and that the 2DLSTM network as access to more source of information and then more learnable concepts through its early 2D-LSTM layers.

All these experiments were made, in priority, to observe the dynamics of results changes between models when some parameters vary. And further tuning of promising architectures should be performed. Nevertheless, we can observe that this 2DLSTM model with feature map depths multiplied by 2, called 2DLSTM-X2 in this paper, obtains the best results of our overall experiments on the difficult datasets that are Maurdor Handwritten, both validation and test, and the historical READ dataset as previously shown in Tables \ref{tab:expeDatasetDev} and \ref{tab:expeDatasetEval}. Moreover, it was not extensively tuned while both the GNN-1DLSTM and the CNN-1DLSTM probably were, respectively by Bluche et al. and by Puigcerver et al.

\begin{table*}[ht]
\begin{center}
\caption{Comparison of GNN-1DLSTM and 2DLSTM models with a varying depth of the feature maps. Results on the French handwritten validation set, no language model (CER, WER).}
\label{tab:expeArchitectureParams}
% \resizebox{\textwidth}{!}{%
\begin{tabular}{lcc}
 & GNN-1DLSTM & 2DLSTM \\
All feature map depths divided by 4 & 32.64\% , 74.15\% & 33.78\% , 75.42\% \\
All feature map depths divided by 2 & 16.30\% , 47.85\% & 14.82\% , 45.87\% \\
Ref feature map depths & 11.80\% , 38.85\% & 10.10\% , 33.86\% \\
All feature map depths multiplied by 2 & 11.09\% , 37.24\% & 8.81\% , 31.35\% \\
\end{tabular}
% }
\end{center}
\end{table*}

Finally, we also compared for the two networks the impact of a various amount of regularization, enforced with Dropout, on the text recognition results. In comparison to the reference, where 4 layers have dropout applied on their outputs during training, we train networks with respectively a low and and a high regularization as we train them with dropout on the outputs of respectively 2 and 7 layers. 

We observe in Table \ref{tab:expeDropout} that, for both networks, the best results are obtained with the reference medium amount of dropout. It shows that the initial amount of dropout were correct for both models and let think that the dynamic of the recognition results with respect to the amount of dropout depends more on the used dataset that on the chosen model.

\begin{table}
\begin{center}
\caption{Comparison of GNN-1DLSTM and 2DLSTM models with a varying amount of dropout regularization. Results on the French handwritten validation set, no language model (CER, WER).}
\label{tab:expeDropout}
% \resizebox{\linewidth}{!}{%
\begin{tabular}{lcc}
 & GNN-1DLSTM & 2DLSTM \\
Dropout small & 18.60\% , 54.07\% & 13.17\% , 42.50\% \\
Dropout medium (Ref) & 11.80\% , 38.85\% & 10.10\% , 33.86\%\\
Dropout large & 12.84\% , 39.98\% & 10.79\% , 35.99\% \\
\end{tabular}
% }
\end{center}
\end{table}

\subsection{Robustness and generalization}
\label{sec:expeRobustness}

We also compared the models on their ability to cope with dataset transfer. For this we trained the networks on the Maurdor handwritten dataset and we tested them on the RIMES validation set. We compare these results in Table \ref{tab:expeCrossBase} with the results of networks trained and tested on the RIMES dataset.
We observe that the results of the models in transfer mode are correlated with the standard results and that none of the tested models generalizes better than the others in a transfer setup. 

\begin{table*}
\begin{center}
\caption{Comparison of the transfer generalization abilities of several models trained separately on the RIMES and Maurdor datasets and evaluated on the validation set of the RIMES dataset (CER, WER).}
\label{tab:expeCrossBase}
\resizebox{\linewidth}{!}{%
\begin{tabular}{llccccc}
 & & CNN & GNN & CNN-1D-LSTM & GNN-1DLSTM & 2DLSTM \\
Trained on RIMES & Tested on RIMES & 16.57\% , 56.55\% & 11.71\% , 43.49\%  & 4.27\% , 16.59\% & 4.00\% , 15.91\% & 3.32\% , 13.24\% \\
Trained on Maurdor & Tested on RIMES & 24.11\% , 70.42\% & 17.20\% , 56.09\% & 6.94\% , 26.20\% & 6.17\% , 24.73\% & 5.13\% , 20.61\% \\
\end{tabular}
 }
\end{center}
\end{table*}

\subsection{Impact of language modeling}
\label{sec:expeLM}

As already mentioned, the LSTM layers, whether one or two dimensional, are very important to get proper results. They enable to share the contextual information between the locations and therefore enhance the performances. Nevertheless, it is known \cite{sabir2017implicit} that they can also learn some kind of language modeling. 

Statistical language models are usually used with the neural networks in order to improve handwriting recognition. That is why we can wonder how adding a language model to our recognition system affects the recognition of our various models that have different recurrent layers.

We use the language models described in Section \ref{sec:languagemodel}. Both character ngram models (5gram, 6gram and 7gram) and word 3gram with various vocabulary size are used. Performances are shown in Table \ref{tab:expeLmMaurdor} for the Maurdor handwritten test set and in Table \ref{tab:expeLmRead} for the Read historical document validation set.

\begin{table*}
\begin{center}
\caption{Comparison of several model results with character or word language modeling applied on top of the network outputs. Results on the French handwritten test set (CER, WER).}
\label{tab:expeLmMaurdor}
\resizebox{\linewidth}{!}{%
\begin{tabular}{lcccccc}
& CNN & CNN-1DLSTM & GNN-1DLSTM & 2DLSTM & 2DLSTM-X2 & CNN-1DLSTM-Puigcerver \\
No LM (best pred) & 33.64\% , 82.40\% & 14.29\% , 44.43\% & 12.58\% , 40.74\% & 10.80\% , 37.12\% & 9.36\% , 33.75\% & 9.80\% , 56.66\% \\ 
Char LM - 5gram & 24.38\% , 63.33\% & 12.42\% , 38.97\% & 10.87\% , 35.32\% & 9.86\% , 33.04\% & 8.38\% , 29.42\% & 8.57\% , 30.93\% \\ 
Char LM - 6gram & 23.23\% , 60.92\% & 11.79\% , 37.51\% & 10.27\% , 33.74\% & 9.48\% , 31.83\% & 8.01\% , 28.06\% & 8.19\% , 29.55\% \\ 
Char LM - 7gram & 22.85\% , 60.01\% & 11.58\% , 36.97\% & 10.06\% , 33.32 \% & 9.26\% , 31.55\% & 7.82\% , 27.44\% & 8.13\% , 29.55\% \\ 
Word LM - 3gram - 25k & 25.69\% , 67.23\% & 13.25\% , 40.14\% & 11.85\% , 37.85\% & 11.62\% , 36.68\% & 9.23\% , 31.25\% & 9.56\% , 32.74\% \\ 
Word LM - 3gram - 50k & 23.73\% , 64.41\% & 11.86\%  , 37.65\% & 10.57\% , 35.28\% & 10.25\% , 33.99\% & 8.31\% , 28.86\% & 8.50\% , 30.13\% \\ 
Word LM - 3gram - 75k & 23.14\% , 63.27\% & 11.34\% , 36.67\% & 10.15\% , 34.25\% & 9.91\% , 32.94\% & 8.15\% , 28.39\% & 8.18\% , 29.41\% \\ 
Word LM - 3gram - 100k & 22.47\% , 62.58\% & 11.01\% , 35.68\% & 9.80\% , 33.38\% & 9.58\% , 32.20\% & 7.96\% , 27.81\% & 7.99\% , 28.84\% \\ 
Word LM - 3gram - 200k & 21.94\% , 61.76\% & 10.74\% , 34.84\% & 9.44\% , 32.39\% & 9.19\% , 31.21\% & 7.74\% , 27.17\% & 7.76\% , 28.12\% \\ 
\end{tabular}
}
\end{center}
\end{table*}

\begin{table*}
\begin{center}
\caption{Comparison of several model results with character or word language modeling applied on top of the network outputs. Results on the historical READ validation set (CER, WER).}
\label{tab:expeLmRead}
\resizebox{\linewidth}{!}{%
\begin{tabular}{lcccccc}
& CNN & CNN-1DLSTM & GNN-1DLSTM & 2DLSTM & 2DLSTM-X2 & CNN-1DLSTM-Puigcerver \\
No LM (best pred) & 26.25\% , 78.84\% & 12.39\% , 41.33\% & 10.40\% , 37.61\% & 7.97\% , 29.48\% & 7.71\% , 29.31\% & 7.83\% , 29.46\% \\
Char LM - 5gram & 10.92\% , 38.75\% & 8.56\%, 29.48\% & 7.68\%  , 28.64\% & 6.08\% , 23.24\% & 5.78\% , 23.37\% & 6.06\% , 24.03\% \\ 
Char LM - 6gram & 9.77\% , 34.87\% & 7.94\% , 27.09\% & 6.94\% , 25.84\% & 5.75\% , 21.91\% & 5.39\% , 21.63\% & 5.68\% , 22.38\% \\ 
Char LM - 7gram & 8.92\% , 31.77\% & 7.60\%, 25.90\% & 6.56\% , 24.54\% & 5.51\% , 20.90\% & 5.23\% , 20.83\% & 5.47\% , 21.45\% \\ 
Word LM - 3gram & 4.88\% , 20.02\% & 4.64\% , 15.35\% & 4.13\% , 14.40\% & 3.30\% , 11.96\% & 3.72\%, 13.73\% & 3.68\% , 13.33\% \\ 
\end{tabular}
}
\end{center}
\end{table*}

For the Maurdor handwritten dataset (cf. Table \ref{tab:expeLmMaurdor}) , for all the models used, both the character and the word language models enable to increase the performances. This is especially true for the larger language models. The best word models tend to give better results than the best character models. Both the model with and without LSTMs get a similar  improvement of about 20\%. The use of a language model and of LSTM layers can therefore be considered complementary.

For the READ dataset, in Table \ref{tab:expeLmRead}, the language models, especially the one made on words, give better results. Similar observations can be made but the CNN network is helped more than the others by the language model.

Even with a language model, the 2DLSTM model shows better performances than the GNN-1DLSTM and the CNN-1DLSTM models of similar sizes. The larger 2DLSTM-X2 model get results similar to the CNN-1DLSTM model from Puigcerver et al. with the biggest word language model and get slightly better results with smaller language models and with character language models.

%Table 2: On Rimes
%\begin{table*}
%\begin{center}
%\caption{Results on Rimes: With or without LM. CER, WER.}
%\label{tab:architecture}
%\resizebox{\linewidth}{!}{%
%\begin{tabular}{lcccccc}
%& CNN & CNN-1DLSTM & GNN-1DLSTM & 2DLSTM & 2DLSTM-X2 & CNN-1DLSTM-Puigcerver \\
%No LM (best pred) &  &  &  &  &  &  \\ 
%Char LM - 5gram &  &  &  &  &  &  \\ 
%Char LM - 6gram &  &  &  &  &  &  \\ 
%Char LM - 7gram &  &  &  &  &  &  \\ 
%Word LM - 3gram - 25k &  &  &  &  &  &  \\ 
%Word LM - 3gram - 50k &  &  &  &  &  &  \\ 
%Word LM - 3gram - 75k &  &  &  &  &  &  \\ 
%Word LM - 3gram - 100k &  &  &  &  &  &  \\ 
%Word LM - 3gram - 200k &  &  &  &  &  &  \\ 
%\end{tabular}
%}
%\end{center}
%\end{table*}

\section{Visualizations}
\label{sec:visualizations}

In previous sections, we discussed the information that the recurrent layers (here the LSTM layers) convey. We said that LSTM layers were important to get contextual information and to avoid noise. We also hypothesized that the 2DLSTM based models were working better on difficult datasets because they were able to better localize the noises thanks to the spatiality of their recurrences. 

\begin{figure}[t]
   \centering
   \includegraphics[width=\linewidth]{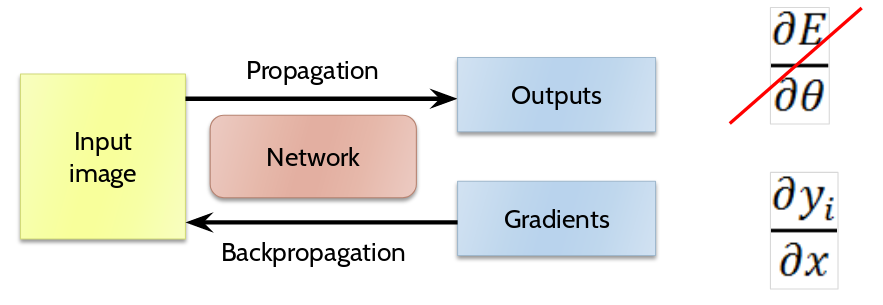}
   \caption{Illustration of the mechanism used to back-propagate the gradients related to a given output back into the input image space.}
   \label{fig:illustrationsVisu}
\end{figure}

This information that is going through the LSTM layers can be visualized by back-propagating some gradients toward the input images space. The gradients follow, backward, the same path the information were transmitted forward. Consequently, we can observe some kind of an attention map, in the input image space, corresponding to the places that were useful to predict a given output. The generic process is illustrated in Figure \ref{fig:illustrationsVisu}.

In order to visualize this attention map, we do the forward pass. Then, we apply a gradient of value 1 for a given output (corresponding to a given character), for all the horizontal positions of the sequence of predictions. No gradient is applied on the other outputs (the other characters). We then back-propagate these gradients in the network without updating the free parameters. We then show the absolute value of this gradient as a map in the input image space. Formally, for all position $x,y$ of the input image $In$ and for a given element $i$ of the output $Out$ , we look for the map that corresponds to: 

\begin{equation}
\forall{x,y} ~ ~ , ~ ~ \frac{\partial Out_{i}}{\partial In_{x,y}} 
\end{equation}

\begin{figure}[t]
   \centering
    CNN: \includegraphics[width=\linewidth]{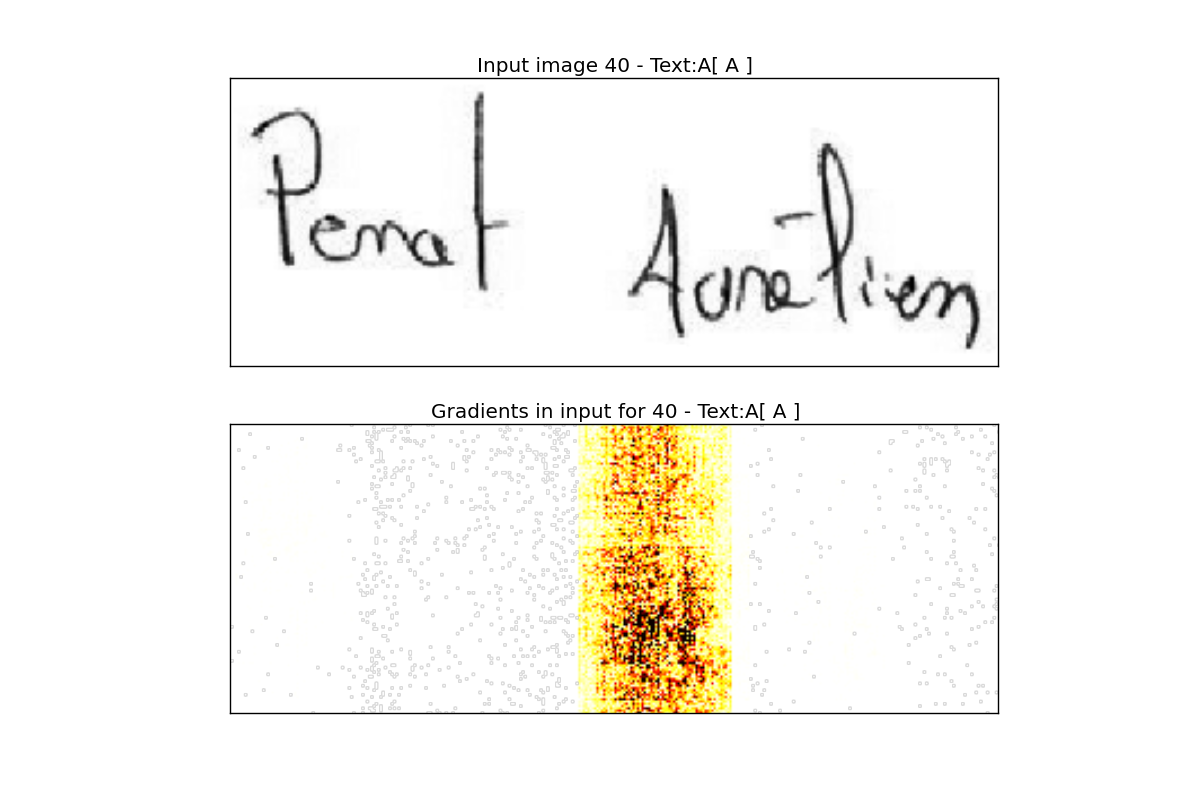}
    GNN-1DLSTM: \includegraphics[width=\linewidth]{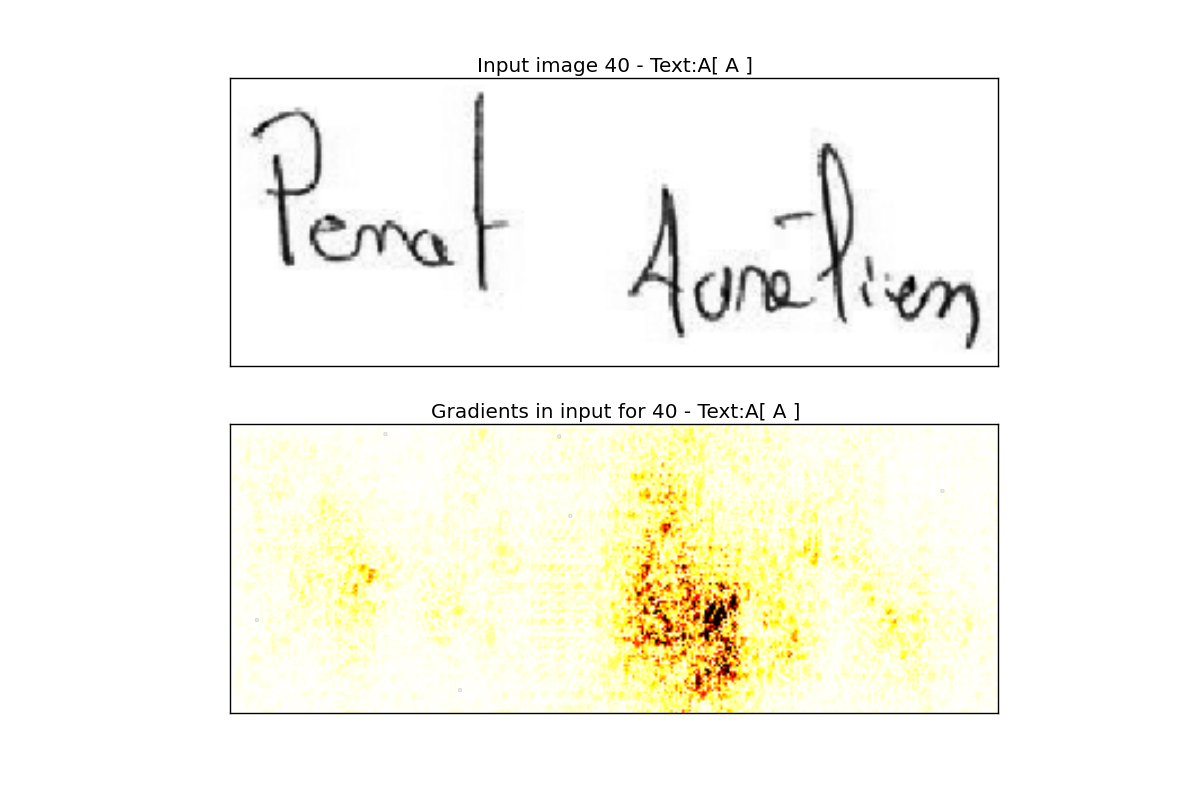}
    2DLSTM: \includegraphics[width=\linewidth]{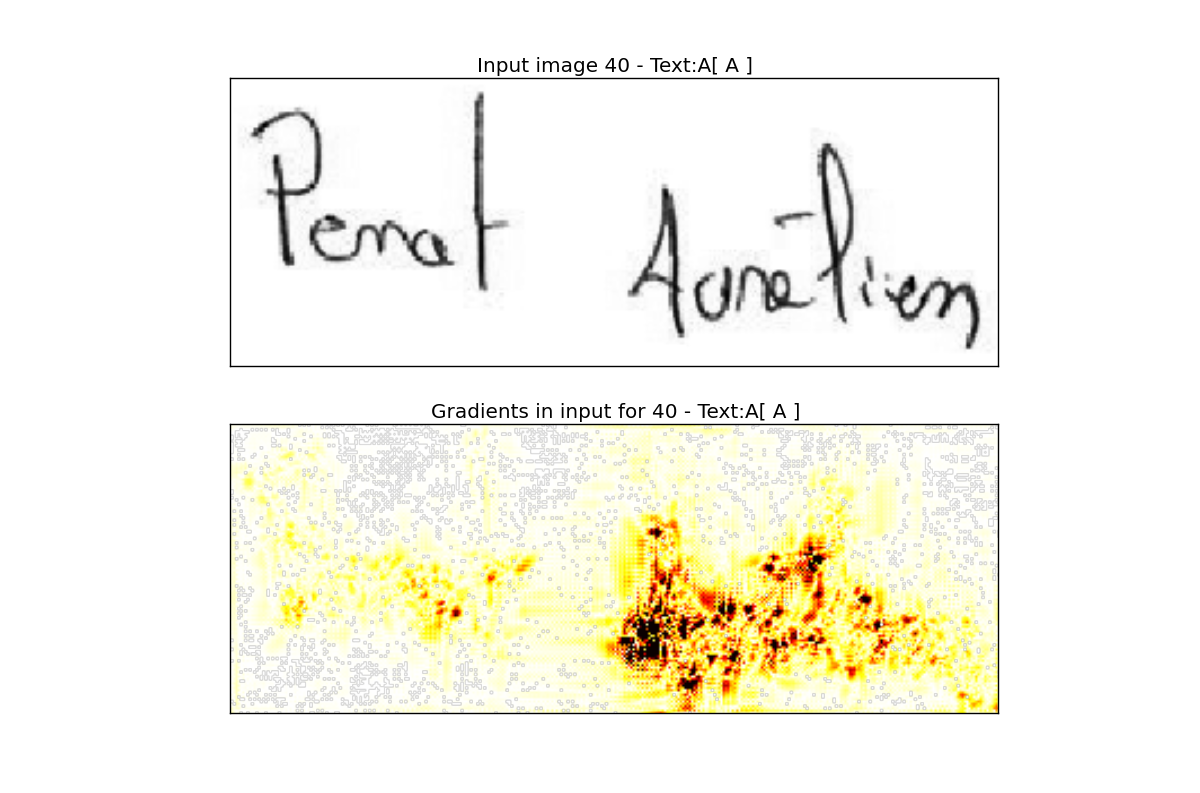}
    %\vspace{1px}
    \\
   \caption{Example, for several different neural network models, of gradients back-propagated in the input image for the outputs corresponding to the letter 'A' .}
   \label{fig:illusGrads1}
\end{figure}

\begin{figure}[t]
   \centering
    CNN: \includegraphics[width=\linewidth]{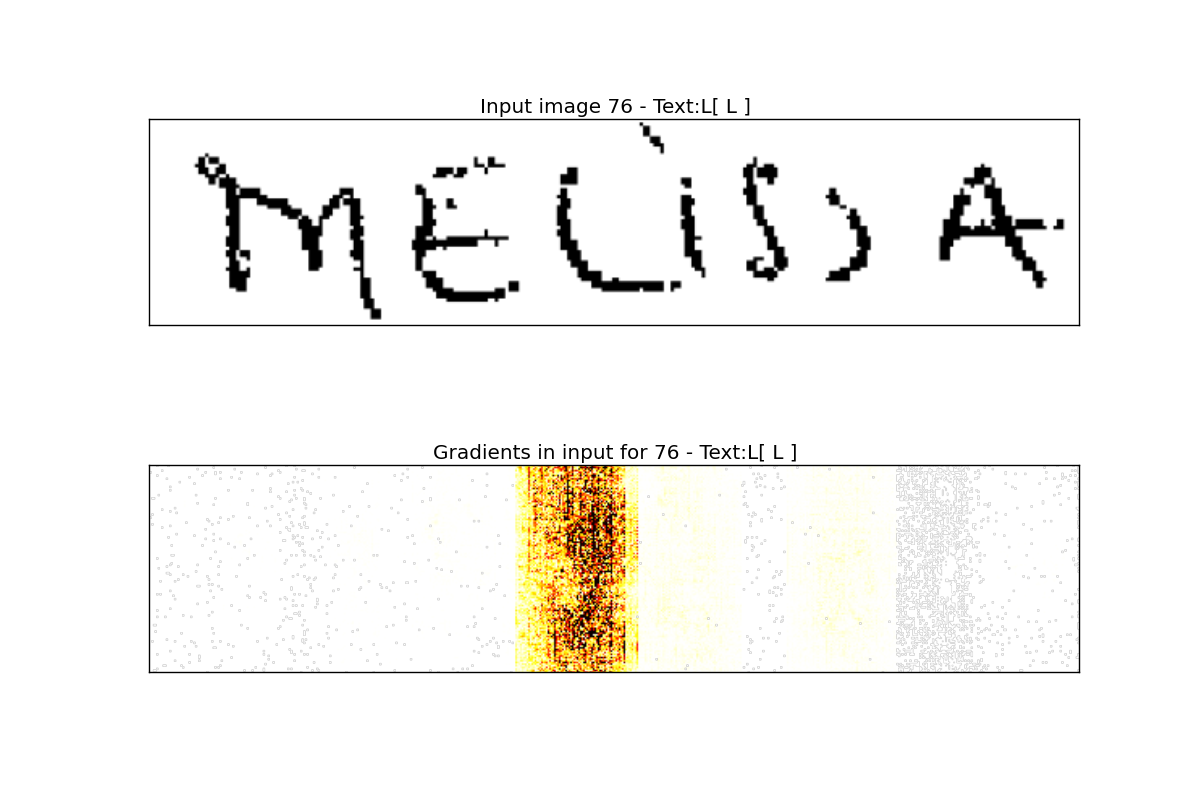}
    GNN-1DLSTM: \includegraphics[width=\linewidth]{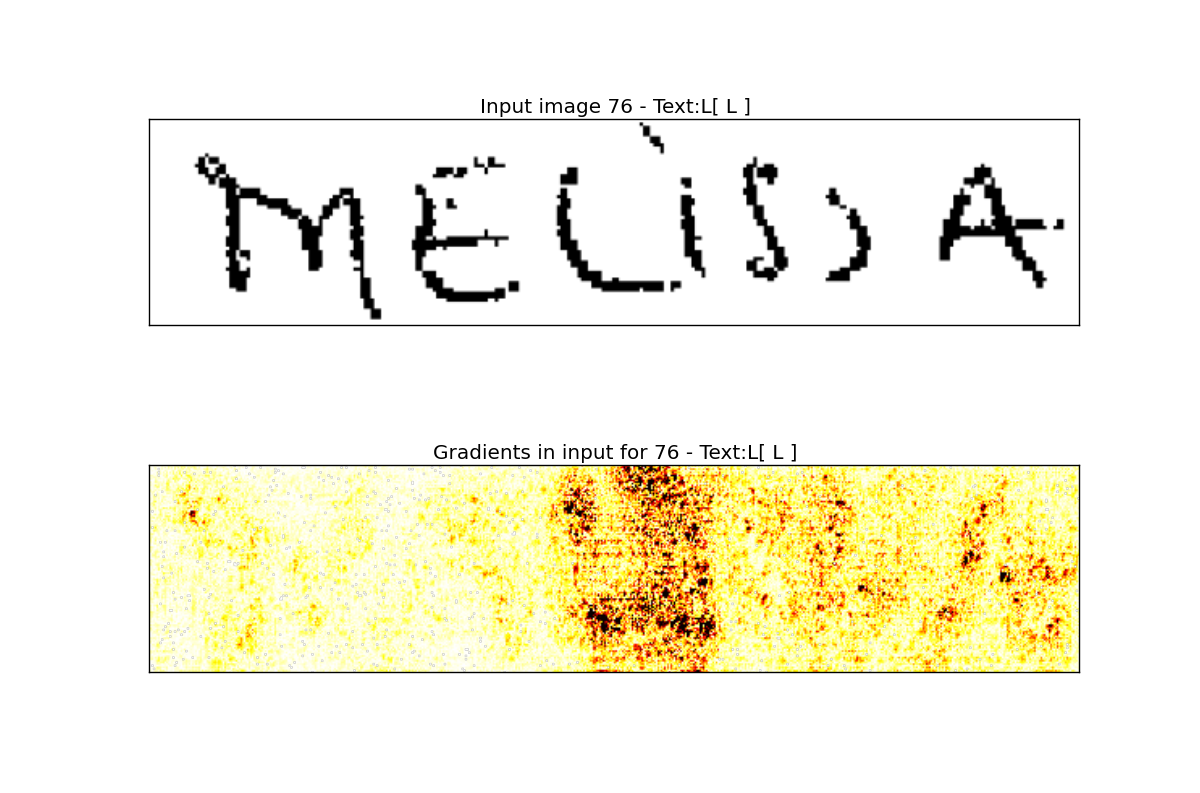}
    2DLSTM: \includegraphics[width=\linewidth]{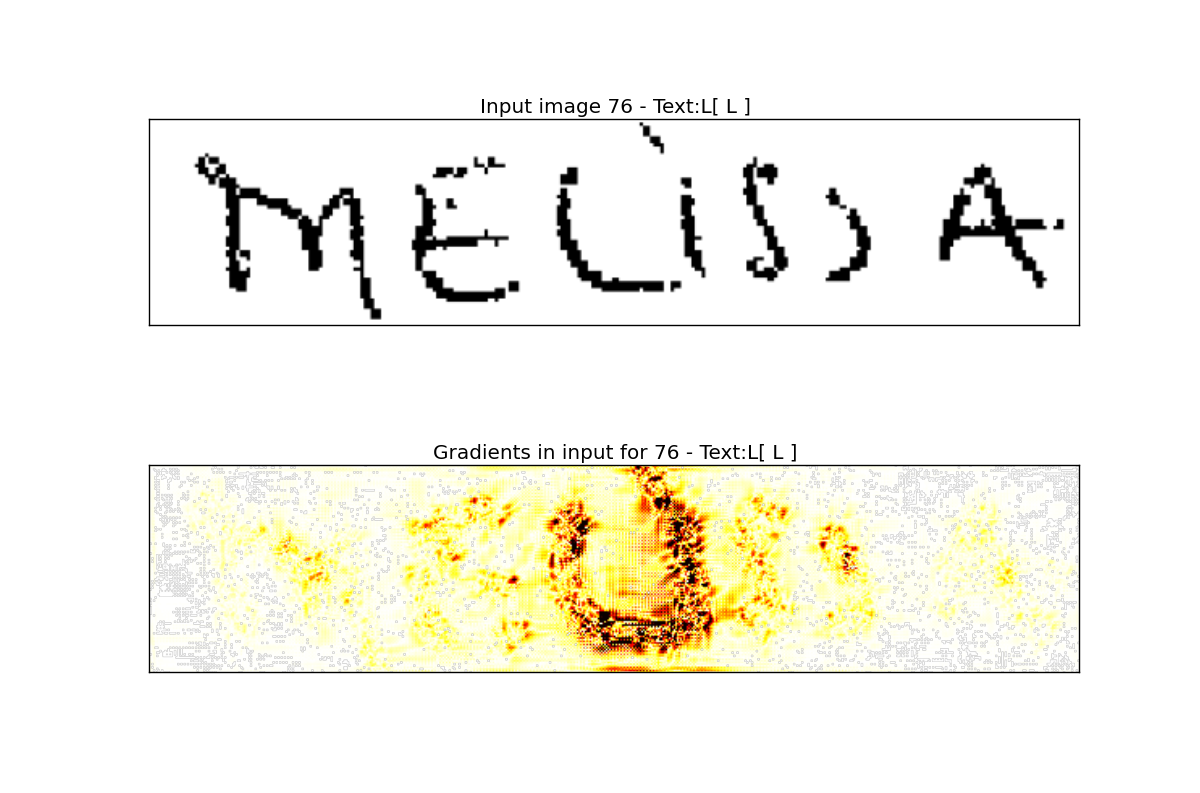}
    %\vspace{1px}
    \\
   \caption{Example, for several different neural network models, of gradients back-propagated in the input image for the outputs corresponding to the letter 'L' .}
   \label{fig:illusGrads2}
\end{figure}

Examples of these attention maps can be visualized for two different images and for three models with different architectures in Figures \ref{fig:illusGrads1} and \ref{fig:illusGrads2}.
 
In the first one, in Figure \ref{fig:illusGrads1}, we observe the gradients of the outputs corresponding to the character 'A'. For the CNN model, the attention is very localized and corresponds mostly to the receptive fields of the outputs that predict this 'A'. We can see that the classification is confident because no other position in the image is likely to predict another 'A'. 

Moreover, because there is no recurrences in this convolutional only network, no attention is put on other places of the image. On the contrary, for the GNN-1DLSTM and 2DLSTM networks, some attention is put outside of these receptive fields. The model will put some attention on the remaining of the word to enhance its confidence that the letter is an 'A'. 

We observe that the attention is sharper for the 2DLSTM model compared to the GNN-1DLSTM one. It can be explained by the fact that this attention can be conveyed at lower layers of the network. 

This sharpness is even more visible on the second image in Figure \ref{fig:illusGrads2} where the networks put attention related to the character 'L'. We can see that the attention is really put on the 'L' itself and on the following 'i' that is crucial to consider in order not to predict another letter (a 'U' for example). For the GNN-1DLSTM network, the attention is more diffuse. 

This possibility offered by 2DLSTMs used on low layers of the networks to precisely locate and identify objects can explain why this system perform slightly better in our experiments on difficult datasets where there is noise and presence of ascendant and descendant from neighboring lines.

\section{Conclusion}
\label{sec:conclusion}

In this work we evaluated several neural network architectures, ranging from purely feed-forward (mainly convolutional layers) to architectures using 1D- and 2D-LSTM recurrent layers. All architectures resort to strided convolutions to reduce the size of the intermediate feature maps, reducing the computational burden. We used two figures of merit to evaluate the performance on text recognition on located lines: character and word error rates over the most probable network output. 
Different number of features were used to evaluate the impact of model complexity over the results; we also varied the amount of dropout to infer the role of regularization for larger models.

To analyze the possible impact of a ``latent'' language model being learned by the architectures using recurrent layers, we also measured the performance when statistical language models are used during the search for the best recognition hypothesis. Word and character-based language models were used to provide a broad range of linguistic constraints over the ``raw'' network outputs.

One of the aims of the study was to provide a fair comparison between the different architectures, while trying to answer the question if 2D-LSTM layers are indeed ``dead'' for text recognition. We also present some visualizations based on back-propagating to the input image space from the gradient of each character in isolation. This results in a kind of ``attention map'', showing the most relevant regions of the input for each recognized character.

Datasets of varying complexity were used in the experiments. Complexity comes from the amount of diversity in writing styles, contents, the presence of noises such as JPEG artifacts, degradations and in some cases, the presence of the ascendants and descendants from the neighboring lines. When the material to be recognized is less complex, such as the case of machine printed text, networks that have less ``modeling power'' (e.g. purely feed-forward) are enough for sufficiently high performance.

Our results show that having LSTM in the networks is essential for the handwriting recognition task. For simple datasets, results do not differ much between 1D and 2D LSTM networks and people can without harm use the more parallelizable 1D architectures. But for more complicated datasets, it seems that the 2D-LSTM are still the state-of-the-art for text recognition. 

Contrarily to what could be expected, adding a language model to networks comprising recurrent layers does improve the performance in a large set of conditions.
We argue that 2D-LSTM can provide the network with sharper ``attention maps'' over the input space of the images, enabling the optimization process to find network parameters that are less sensitive to the different noises in the image.

%\begin{acknowledgements}
%\end{acknowledgements}

\bibliographystyle{spmpsci}      % mathematics and physical sciences
\bibliography{refs}   % name your BibTeX data base

% \begin{thebibliography}{}
% %
% % and use \bibitem to create references. Consult the Instructions
% % for authors for reference list style.
% % Format for Journal Reference
% Author, Article title, Journal, Volume, page numbers (year)
% Author, Book title, page numbers. Publisher, place (year)
% \end{thebibliography}

\end{document}